\documentclass[journal]{IEEEtran}
\usepackage{multirow}
\usepackage{amsmath}
\usepackage{booktabs}

\usepackage[dvipsnames]{xcolor}
\usepackage{multirow}
\usepackage[shortlabels]{enumitem}
\newtheorem{definition}{Definition}

\ifCLASSINFOpdf
   \usepackage[pdftex]{graphicx}
\else
\fi
\begin{document}
%
\title{GCCN: Global Context Convolutional Network}
%
%
%

\author{Ali~Hamdi,
        Flora~Salim,
        and~Du Yong~Kim
\thanks{Ali Hamdi and Flora Salim are with the School of Computing Technologies, RMIT University, Australia, ali.ali@rmit.edu.au, flora.salim@rmit.edu.au.}
\thanks{Du Yong Kim is with the School of Engineering, RMIT University, Australia, duyong.kim@rmit.edu.au.}
}

%
%

\markboth{Ali Hamdi \MakeLowercase{\textit{et al.}}: GCCN: Global Context Convolutional Network}
{Ali Hamdi \MakeLowercase{\textit{et al.}}: GCCN: Global Context Convolutional Network}
%



\maketitle

\begin{abstract}
In this paper, we propose Global Context Convolutional Network (GCCN) for visual recognition. GCCN computes global features representing contextual information across image patches. These global contextual features are defined as local maxima pixels with high visual sharpness in each patch. These feature are then concatenated and utilised to augment the convolutional features. The learnt feature vector is normalised using the global context features using Frobenius norm. This straightforward approach achieves high accuracy in compassion to the state-of-the-art methods with $94.6\%$ and $95.41\%$ on CIFAR-10 and STL-10 datasets, respectively. To explore potential impact of GCCN on other visual representation tasks, we implemented GCCN as a based model to few-shot image classification. We learn metric distances between the augmented feature vectors and their prototypes representations, similar to Prototypical and Matching Networks. GCCN outperforms state-of-the-art few-shot learning methods achieving $99.9\%$, $84.8\%$ and $80.74\%$ on Omniglot, MiniImageNet and CUB-200, respectively. GCCN has significantly improved on the accuracy of state-of-the-art prototypical and matching networks by up to $30\%$ in different few-shot learning scenarios.
\end{abstract}


%
\IEEEpeerreviewmaketitle

\section{Introduction}
%
%
%
%
\IEEEPARstart{C}{onvolutional} 
 Neural Networks (CNNs) have been employed to learn local image features through an isotropic mechanism of their receptive fields \cite{luo2016understanding}. Typically, CNNs struggle to deal with global contextual features. Therefore, conventional CNNs are unable to capture useful structural information and deal with diverse backgrounds and unrepresentative regions. 
Recent studies in CNN proposed to various method to find the optimal receptive field. Despite this, they implement conventional architecture search methods in coarse search spaces. This mechanism loses important fine-grained inner structures \cite{Liu_2021_CVPR}. The flexibility of CNN offers an excellent opportunity to solve this problem by designing  different architectures \cite{szegedy2015going,he2016deep,huang2017densely,tan2019efficientnet}. These models went deeper and wider with convolution neural networks. 
These models are trained on large scale image datasets, e.g., ImageNet, and employed different data augmentation techniques to achieve high accuracy and solve overfitting issues \cite{masi2016we}. Increasing the network depth and width is hard to train due to the vanishing gradient problem.
Moreover, patch-based models, especially with Graph Convolutional Networks (GCN), have been introduced to capture global visual features \cite{zhou2018graph,gao2019graph,hamdi2020flexgrid2vec}. However, recent GCN models typically suffer from the increasing size and complexity of the network parameters and computations.

\begin{figure*}[!t]
\begin{center}
   \includegraphics[width=0.99\linewidth]{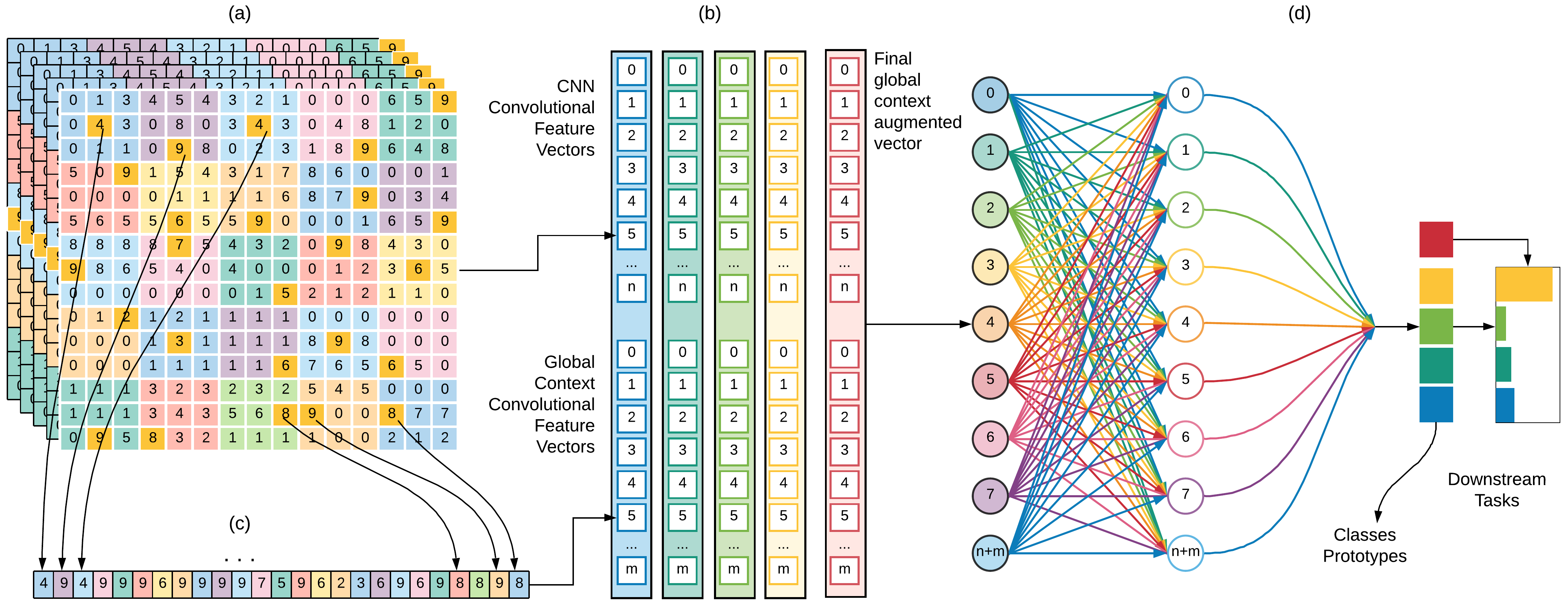}
\end{center}
   \caption{GCCN extracts global context information from CNN feature maps. a) Convolutional feature maps are utilised to produce feature embedding and global context features. c) Global context feature vector extraction from image patches. b) Feature vector augmentation and normalisation. d) A head model for few-shot learning computes the member class distributions based on the metric distance measures.}
   \label{GCCN2}
\end{figure*}

In this paper, we propose a novel patch-based method to compute global context features to capture significant complex structures in images. The proposed Global Context Convolutional Network (GCCN) offers a powerful yet straightforward approach to augment and normalise the classical CNN feature vectors. 
GCCN computes features of local maxima of image patches based on CNN feature maps, as visualised in Fig. \ref{GCCN2}. Local maxima convolutional features represent pixels with high visual sharpness after convolution and pooling operations. Therefore, connecting local maxima features from different image regions tend to have a discriminative feature vector.
GCCN shows significant accuracy in image classification.

We explore the potential impact of using GCCN to achieve accurate few-shot learning, as in Fig. \ref{GCCN1}. Learning from a few samples is a challenging task in visual representation learning. Current methods do not offer satisfactory solutions for few-shot learning~\cite{vinyals2016matching}. Naive methods that depend on retraining the model on the new data is extremely overfitting \cite{snell2017prototypical}. The overfitting problem leads to limited scalability to learn new classes and poor applicability to fit new unseen or rare examples \cite{sung2018learning}. Existing method that overcome overfitting such as batch and layer normalisation \cite{ioffe2015batch,ba2016layer}, usually, fail in few-shot \cite{antoniou2017data}. 
GCCN produces discriminative visual feature with in-depth attention to local and global contexts. The learnt feature space is utilised to compute the centroid of each class similar to \cite{snell2017prototypical,mensink2013distance,rippel2015metric}. The proposed \textit{GCCN} outperforms recent works such as VAMPIRE (WACV, 2020) \cite{nguyen2020uncertainty}, APL (ICLR, 2019) \cite{ramalho2019adaptive}, SImPa (TPAMI, 2020) \cite{nguyen2020pac}, LaplacianShot (ICML, 2020) \cite{ziko2020laplacian}, and Hyperbolic ProtoNets (CVPR, 2020) \cite{khrulkov2020hyperbolic}. 
{Figure \ref{GCCN2} shows the components of the proposed GCCN. We extract global contextual features from convolutional maps. First, GCCN uses these maps to compute visual embedding and selects global context features from image patches. We then apply feature vector augmentation and normalisation. A fully connected layer uses GCCN final vectors to perform image classification. On the other hand, a head model for few-shot learning computes the member class distributions based on the metric distance measures, as shown in Fig. \ref{GCCN2} (d).}

The proposed GCCN is simple yet efficient enhancing CNN accuracy with computing small size context feature vectors. GCCN is also designed to be flexibly applied on different CNN architectures. The main contributions are as follows:
\begin{itemize}
    \item A novel method to augment and normalise CNN features with global context information. GCCN tends to produce useful feature embeddings with attention to both local and global image structures.
    \item An implementation of the proposed \textit{GCCN} as a base model to the state-of-the-art prototypical and matching networks. \textit{GCCN} improves their accuracy by up to $30\%$ in a variety of few-shot learning benchmarks.
    \item A comparative study on image classification and few-shot learning tasks with well-known baseline architectures and state-of-the-art on seven benchmark datasets, namely CIFAR-10, CIFAR-100, STL-10, SVHN, Omniglot, MiniImageNet and CUB-200.
\end{itemize}

{
The rest of this paper is organised as follows. Section \ref{rw} reviews the related literature of image classification and few-shot learning. Sections \ref{gccn} introduces the proposed GCCN and its components. Section \ref{results} the experimental work and discusses the accuracy of GCCN compared to baseline and state-of-the-art methods.} 

\section{Related Work}\label{rw}
\paragraph{Image Classification.}
Visual representation learning has advanced image classification tasks. Recent model offer multiple deep learning method to compute representative visual features such as ResNets \cite{he2016deep,huang2016deep,he2016identity}, DenseNet \cite{huang2017densely}, and Efficient-Net \cite{tan2019efficientnet}. Such models offer alternative methods to compute better convolutional features in comparison to the classical CNN. They proposed new architecture designs based on CNN depth (number of layers) and width (number of neurons in each layer). Recent models used to be pre-trained on the ImageNet dataset and accompanied by image augmentation techniques. Recent methods have also introduced useful loss functions to learn discriminative feature space through effective gradients. Deep networks, such as ResNet and Efficient-Net, have achieved state-of-the-art accuracy in various image classification tasks. They have also inspired multiple recent studies to propose new CNN based visual representation learning methods for different vision applications. Although they offer deep, wide, and effective architectures, they are still limited by the complex visual structures, available resources. 

\paragraph{Few-Shot Learning.}
Few-shot learning has different research directions, including metric learning \cite{vinyals2016matching,snell2017prototypical,sung2018learning}, transfer learning \cite{chen2019closer}, meta-learning \cite{finn2017model}, and data augmentation approaches \cite{antoniou2017data,wang2018low}. Matching networks \cite{vinyals2016matching} trains attention memory-based classifier over episodes of support and query sets. Matching networks use LSTM to update the few-shot classifier at each episode to generalise to the test set. Although this approach is complex, it still relies on distance similarity metrics. Sachin et al., \cite{Sachin2017Optimization} employed this episode strategy to learn meta-learning for few-shot learning.  Prototypical networks \cite{snell2017prototypical} are proposed to overcome the few-shot overfitting problem. They are designed to learn the class centroids or prototypes in the feature space. These prototypes are computed as the means of non-linear CNN based feature embeddings. Prototypical networks have been employed in multiple recent works such as Hyperbolic ProtoNets \cite{khrulkov2020hyperbolic}. Their simple design enables further research to extend on them. In~\cite{wang2018low}, both Prototypical and Matching networks are hallucinated with image augmentation via generative model. However, data augmentation methods offer limited reasonable images variations. There is a need for new research to develop much broader augmentations.

\section{Global Context Convolutional Network}\label{gccn}
This paper proposes a novel method to learn global and local visual features through a straightforward yet effective approach.  
{
In this Section, we introduce the proposed architecture for two visual recognition tasks, namely, image classification and few-shot learning as visualised in Figures (\ref{GCCN2} and \ref{GCCN1}, respectively. 
}
GCCN computes attention features from different image regions without the need for complex, wide or large architectures. The proposed \textit{GCCN} combines the CNN feature vector with global context features. Inspired by the metric learning of distances of the query and the support centroids \cite{snell2017prototypical}, our proposed methodology enhances few-shot learning by augmenting and normalising the CNN feature vectors with important global information. Our main contribution enables informative convolution vector embedding representations of both the local appearance and global context of an image. This simple yet powerful representation tends to be helpful for both image classification and metric learning. Fig. \ref{GCCN1} and \ref{GCCN2} show the input images and the different components of the proposed algorithm. 

\subsection{GCCN for Visual Representation learning}
We propose a novel base model for visual feature extraction. This model augments the CNN feature embeddings with global context information to overcome the CNN limitation of ignoring global structural features due to the local receptive fields. 
{
We compute the global contextual feature vector as a concatenation of the local maxima from different region on the image feature maps. This global vector is utilised to augment and normalise the conventional CNN embedding vector. The proposed \textit{GCCN} is an algorithmic framework that includes multiple components. Fig. \ref{GCCN2} describe the GCCN architecture, as follows: 
\begin{enumerate}[(a)]
    \item Dividing CNN feature maps into non-overlapping patches. 
    \item Extracting CNN classical feature vectors.
    \item Computing CNN based global feature vectors, as defined in Definition \ref{GC}.
    \item Preforming vector augmentation and normalisation between the convolutional and global context features. 
\end{enumerate}
}

{
\begin{definition}\label{GC}
\textbf{Global Context (GC)} features vector is defined as a set of key visual points computed based on CNN feature maps. These key points are selected as local maxima after the convolutional and max-pooling operations.
\end{definition}
}

First, we extract the CNN feature vectors for both $S$ and $Q$ images. Then, we extract global context convolution feature vectors for each image. We divide the convolution feature map into small equal patches via a sliding window. The CNN features maps of an image $I$ computed by a filter kernel $K$ as follows:
\begin{equation}
    conv(I,K)_{x,y} = \sum^h_{i=1} \sum^w_{j=1} \sum^c_{k=1} K_{i,j,k} I_{x+i-1,y+j-1,k}
\end{equation}
where $x$ and $y$ denote the coordinates of the image and $h$, $w$ and $c$ are the height, width and image channels. 
{
This feature map is divided into small patches, and the local maxima i.e., pixels with maximum values, are selected to form the global context feature vector $GC$, as defined in Definition \ref{GC}. The selected features are concatenated in one vector.
\begin{equation}
    GC = \sum_{W_i \in W} max(W_i)
\end{equation}
where $W$ is a set of patches of the feature map. The output $GC$ is concatenated with the CNN feature vector to present the feature vector augmentation, as in Eq. \ref{GCconcat}.
\begin{equation}\label{GCconcat}
GCCN(I) = conv(I) \bigoplus GC
\end{equation}
This feature extraction mechanism has effectively enabled the CNN based model to augment its local features with a set of global context informative features.}

{
We propose to utilise the extracted GCCN features to normalise the CNN features. The Euclidean (also called Frobenius) norm is computed as Eq. \ref{norm}.
\begin{equation}\label{norm}
\|V\|_{F}=\left[\sum_{i, j} \operatorname{abs}\left(a_{i, j}\right)^{2}\right]^{1 / 2}
\end{equation}
where $F$ denotes the Frobenius norm, $V$ is the feature vector and $a_{i, j}$ is an element in $V$. This vector normalisation method calculates the square root of the sum of absolute squares of the matrix elements. The output of the norm process is utilised to normalise the original or augmented CNN vector, as in Eq. \ref{GCnorm}.
\begin{equation}\label{GCnorm}
GCCN(I) = \frac{conv(I) \bigoplus GC}{\|GC\|_{F}} 
\end{equation}
}

We provide extensive experimental work to compare these different setups on four image classification datasets, including CIFAR-10, CIFAR-100, STL-10 and SVHN. 
Next, we use the augmented feature vectors to compute each class's prototype or utilise it by the matching networks.

\subsection{GCCN for Few-shot Learning}
{
Deep learning methods require large datasets to be trained and fitted across a large number of parameters. The available training datasets are limited in many realistic scenarios, leading to poor model generalisation on other test data. Few-shot learning enables classification models to be trained on a few samples. It can be one-shot or more based on how many samples per class. It offers data-efficient learning and a more efficient approach with regards to fine-tuning and model adaptation \cite{antoniou2019how}. Learning from a few samples is a challenging task in visual representation learning. Current methods do not offer satisfactory solutions for few-shot learning~\cite{vinyals2016matching}. Naive methods that depend on retraining the model on the new data is extremely overfitting \cite{snell2017prototypical}. The overfitting problem leads to limited scalability to learn new classes and poor applicability to fit new unseen or rare examples \cite{sung2018learning}. Existing method that overcome overfitting such as batch and layer normalisation \cite{ioffe2015batch,ba2016layer}, usually, fail in few-shot \cite{antoniou2017data}. 
}

\begin{figure*}[ht]
\begin{center}
   \includegraphics[width=0.99\linewidth]{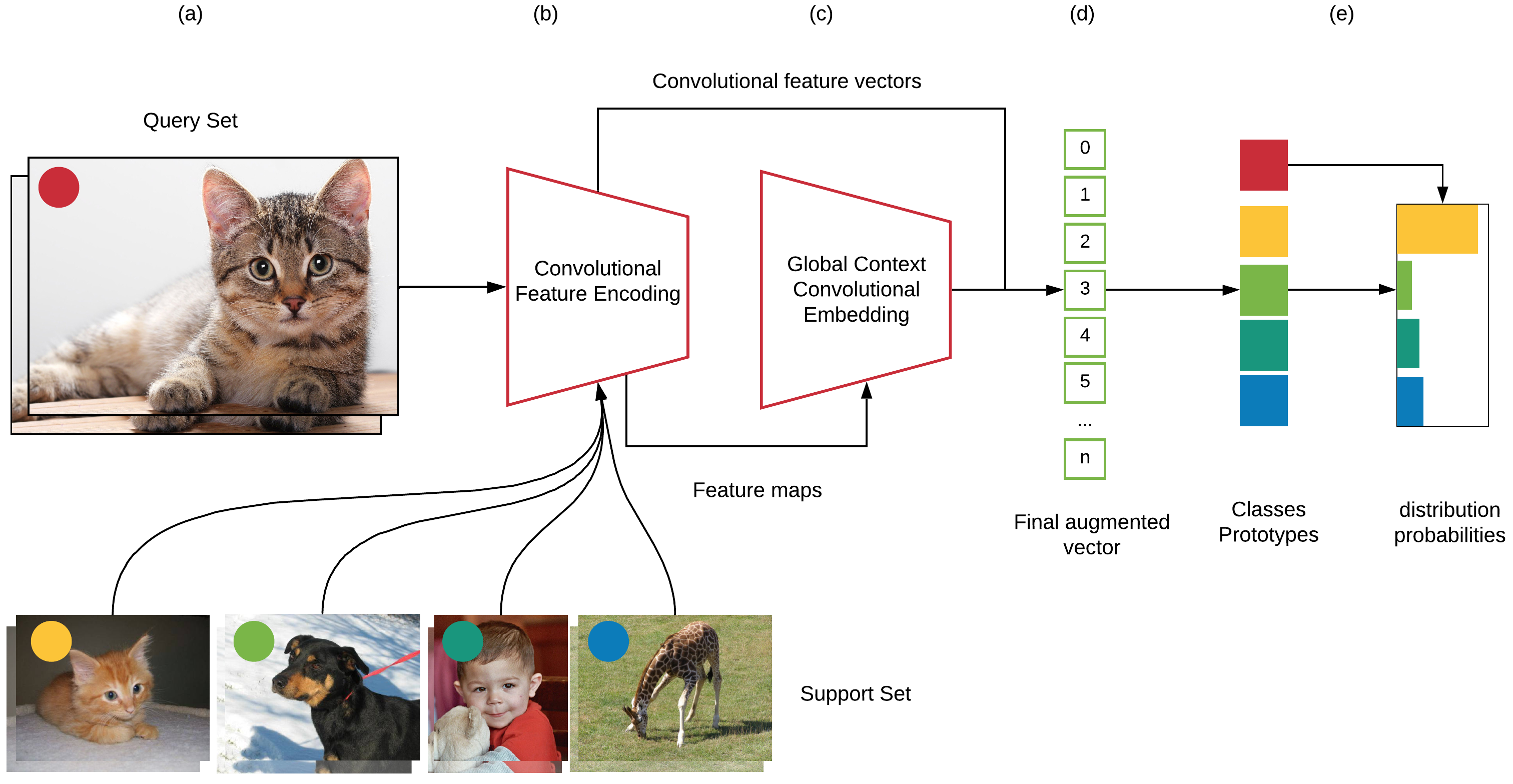}
\end{center}
   \caption{
   {
   The proposed architecture: global context convolutional neural networks for few-shot learning. a) Support $S$ and query $Q$ sets. b) Extracting convolution feature. c) global context feature vector based on the CNN feature maps. d) CNN feature vector augmentation with the global feature vector. e) A head model for few-shot learning. Prototypical networks are used to compute the class prototypes.}\}}
   \label{GCCN1}
\end{figure*}

We define few-shot learning as mapping a support set of small $k$ samples $S = \{(x_i, y_i)\}^k_{i=1}$ to a classifier $c_S(\hat x)$. That classifier is given a test image $\hat x$ to find the probability distribution over the $\hat y$ output labels. Fig. \ref{GCCN1} shows the main components of the proposed few-shot image classification architecture using \textit{GCCN}. The figure incorporates four main sections: the input query and support image sets, the extraction of the CNN feature, the vector augmentation and the learning process of the probability distribution between the query vector and the support prototypes.

\paragraph{Prototypical Networks.}
The output GCCN feature vectors are fed into a few-shot learning metric-based model. We use state-of-the-art prototypical networks for their simplicity and high accuracy. Prototypes are defined as the estimated mean of each class of vectors. These vectors are computed over the support set $S$ to measure the distance between them and the vector of the query image $Q$. The latter is assigned to the closest class mean or prototype. 
The works in \cite{mensink2013distance,rippel2015metric} use multiple prototypes per class. However, having more than one prototype in each class requires a portioning function to group each class of support points. In this paper, we only select one prototype per class, similar to \cite{snell2017prototypical}.
The embedded query vector is classified through softmax over distances to the class prototypes.
\begin{equation}
  p_\phi(y=k|x) = \frac{e^{(-d(f\phi(X), \mu_k))}}{\sum_{k'} e^{(-d(f\phi(x), \mu_{k'})}}
\end{equation}
where $d$ is a distance function, $f\phi$ a feature vector embedding function and $\mu_k$ is the prototype as in Eq. \ref{mu}.
\begin{equation}\label{mu}
\mathbf{\mu}_{k}=\frac{1}{\left|S_{k}\right|} \sum_{\left(\mathbf{x}_{i}, y_{i}\right) \in S_{k}} f_{\phi}\left(\mathbf{x}_{i}\right)
\end{equation}

\paragraph{Matching Networks.}
We have also implemented \textit{GCCN} under the state-of-the-art matching networks. They utilise a weighted nearest-neighbour classifier over embedding space. $S$ is modelled in a sequence, and $Q$ is embedded in it by a bidirectional long short-term memory (LSTM). This work defines the one-shot learning task as mapping a support set of small $k$ samples $S = \{(x_i, y_i)\}^k_{i=1}$ to a classifier $c_S(\hat x)$. That classifier is given a test image $\hat x$. It aims to find the probability distribution over the $\hat y$ output labels. The mapping is defined as follows:
\begin{equation}\label{P}
  S \rightarrow c_S(\hat x) \Longleftrightarrow P(\hat y|\hat x, S') = \sum^k_{i=1} a(\hat x, x_i)y_i
\end{equation}
where $P$ is parameterised by a neural network to make the label $\hat y$ prediction and $a$ is an attention mechanism through a $X \times X$ kernel density estimation. The $P$ in Eq. \ref{P} depends on $a$, the attention mechanism that fully controls the classifier. This is simply computed using the softmax over cosine distance $c$, as in Eq. \ref{a}.
\begin{equation}\label{a}
  a(\hat x,x_i) = \frac{e^{c(f(\hat x),g(x_i))}}{\sum^k_{j=1} e^{c(f(\hat x),g(x_j))}}
\end{equation}
where $f$ and $g$ are embedding functions through neural networks to embed $\hat x$ and $x_i$. 

\paragraph{Distance and Similarity Metrics}
We tested the $Euclidean$ distance method as follows:
\begin{equation}\label{euclidean}
  E(p,q) = \sqrt {\sum _{i=1}^{n}  \left( q_{i}-p_{i}\right)^2}
\end{equation}
where $p$ and $q$ are the support/prototype and query vectors. The distance function $E$ is Pythagorean formula and $q$ and $p$ are normalised by the Euclidean norm $L_2$.
\begin{equation}
  ||p|| = \sqrt{p^2_1+p^2_2+...+p^2_n} = \sqrt{p\cdot p}
\end{equation}
Eq. \ref{euclidean} can now be written as follows:
\begin{equation}
  ||q-p|| \sqrt{(q-p)\cdot (q-p)} = \sqrt{||p||^2+||q||^2-2p\cdot q}
\end{equation}
We have also tested the cosine similarity method, as in Eq. \ref{cos}.
\begin{equation}\label{cos}
  \cos ({\bf p},{\bf q})= {{\bf p} {\bf q} \over \|{\bf p}\| \|{\bf q}\|} = \frac{ \sum_{i=1}^{n}{{\bf p}_i{\bf q}_i} }{ \sqrt{\sum_{i=1}^{n}{({\bf p}_i)^2}} \sqrt{\sum_{i=1}^{n}{({\bf q}_i)^2}} }
\end{equation}
The $d$ distance function in the above-mentioned head models can be $E$ or $cos$. We introduce an extensive experimental discussion on both distance methods in the next section. 

\section{Experimental Results}\label{results}

\subsection{Image Classification}
GCCN offers a novel approach to capture global contextual attentions to support the conventional convolutional features. We evaluate the GCCN as a visual representation method for image classification. We utilised four benchmark datasets, as follows:
\begin{itemize}
  \item CIFAR-10 \cite{krizhevsky2009learning} includes of $60,000$ $32 \times 32$ images. They are divided into $50,000$ for training and $10,000$ for testing. CIFAR-10 dataset has $10$ categories, such as dog, cat and bird. The categories are mutually exclusive with no overlapping. 
    \item CIFAR-100 is similar to the CIFAR-10, having $100$ classes containing $600$ images each.
    \item STL-10 \cite{coates2011analysis} has $10$ classes with $5,000$ images for training and $8,000$ for testing. The images are in the size of $96 \times 96$ and are acquired from the ImageNet dataset. 
    \item SVHN contains $630,420$ house number images of a size $32 \times 32$ pixels. The SVHN official split contains $73,257$ and $26,032$ images for training and testing, respectively.
\end{itemize}

We tested the GCCN using ResNet-50 and Efficient-Net. GCCN is implemented over one convolutional block. We designed the convolutional block to return one feature map after the max pooling. This feature map is utilised to compute the global context vectors. Thus, GCCN can be repeated after each convolutional process over the output pooled feature maps. We tested GCCN with one, two and three layers to explore, as listed in Table \ref{gccn_comp}.

\paragraph{CIFAR-10.}
Table \ref{CIFAR-10} lists the benchmark results of using GCCN on the CIFAR-10 dataset. GCCN has introduced high accuracy with $94.6\%$. GCCN outperforms state-of-the-art models such as DeepInfoMax \cite{hjelm2018learning}, which has 75.57\%. DeepInfoMax is a patch-based approach similar to our GCCN. Although DeepInfoMax has a complex architecture, GCCN achieved higher accuracy. Other recent methods are also outperformed by GCCN, such as 
ANODE \cite{NIPS2019_8577} (NeurIPS, 2019), CLS-GAN \cite{qi2020loss} (IJCV, 2020) and Mish \cite{misra2020mish} (BMVC, 2020).

\begin{table}[!ht]
\small
\begin{center}
\caption{Classification accuracy (top 1) results on CIFAR-10.}\label{CIFAR-10}
\small
\begin{tabular}{|p{5.5cm}|l|}
\hline
Model & Test Accuracy\\ \hline
ANODE \cite{NIPS2019_8577} & 60.6\% \\ \hline 
DeepInfoMax (infoNCE) \cite{hjelm2018learning} & 75.57\%\\ \hline 
DenseNet \cite{huang2017densely} & 77.79\% \\ \hline 
DCGAN \cite{radford2015unsupervised} & 82.8\% \\ \hline
Baikal \cite{gonzalez2020improved} & 84.53\% \\ \hline 
Scat + FC \cite{oyallon2017scaling} & 84.7\% \\ \hline
CapsNet \cite{sabour2017dynamic} & 89.4\% \\ \hline 
MP \cite{hendrycks2016baseline} & 89.07\%\\ \hline
ResNet-34 \cite{he2016deep} & 89.56\% \\ \hline
APAC \cite{sato2015apac} & 89.70\% \\ \hline

MIM \cite{liao2016importance} & 91.5\% \\ \hline 
CLS-GAN \cite{qi2020loss} & 91.7\% \\ \hline 
DSN \cite{lee2015deeply} & 91.8\% \\ \hline 
BinaryConnect \cite{courbariaux2015binaryconnect} & 91.7\% \\ \hline 
Mish \cite{misra2020mish} & 92.20\% \\ \hline 
\textbf{GCCN (ours)} & \textbf{94.6\%} \\ \hline
\end{tabular}
\end{center}
\end{table}

\paragraph{CIFAR-100.} 
Table \ref{CIFAR-100} shows experiment results comparing GCCN to state-of-the-art image classification methods using the CIFAR-100 dataset. We utilised ResNet-50 as a base model to compute the CNN feature within the GCCN. GCCN improves the ResNet-50 accuracy from $64.06\%$ to $79.77\%$. GCCN also outperforms deeper versions of ResNet, such as the ResNet-1001, which achieves $77.3\%$. In the same fashion, GCCN outperforms recent state-of-the-art methods, such as MixMatch \cite{NEURIPS2019_MixMatch} with 74.10\%, Mish \cite{misra2020mish} with 74.41\% and DIANet \cite{huang2020dianet} with 76.98\% .

\begin{table}[!ht]
\small
\begin{center}
\caption{Classification accuracy (top 1) results on CIFAR-100.}\label{CIFAR-100}
\small
\begin{tabular}{|p{5.5cm}|l|}
\hline
Model & Test Accuracy\\ \hline
DSN \cite{lee2015deeply} & 65.4\% \\ \hline 
ResNet-50  \cite{he2016identity} & 67.06\% \\ \hline 
MIM \cite{liao2016importance} & 70.8\% \\ \hline 
MixMatch \cite{NEURIPS2019_MixMatch} & 74.10\% \\ \hline 
Mish \cite{misra2020mish} & 74.41\% \\ \hline 
Stochastic Depth \cite{huang2016deep} & 75.42\% \\ \hline 
Exponential Linear Units \cite{clevert2015fast} & 75.7\% \\ \hline 
DIANet \cite{huang2020dianet} & 76.98\% \\ \hline 
Evolution \cite{real2017large} & 77\% \\ \hline 
ResNet-1001 \cite{he2016identity} & 77.3\% \\ \hline 
\textbf{GCCN} & \textbf{79.77\%} \\ \hline
\end{tabular}
\end{center}
\end{table}

\paragraph{STL-10}
Table \ref{STL-10} lists benchmark results of using the STL-10 dataset. GCCN achieves $95.41\%$ accuracy outperforming state-of-the-art methods. It has better accuracy than different versions of DeepInfoMax, FixMatch and NSGANetV2. ResNet accuracy also improved from $82.66\%$ to $95.41\%$  using the proposed GCCN.

\begin{table}[!ht]
\small
\caption{Classification accuracy (top 1) results on STL-10 dataset.}\label{STL-10}
\centering
\begin{tabular}{|p{6cm}|l|}
\hline
Model & Test Accuracy  \\ \hline
DeepInfoMax (JSD) \cite{hjelm2018learning} & 65.93\% \\ \hline 
DeepInfoMax (infoNCE) \cite{hjelm2018learning} & 67.08\% \\ \hline 
ResNet \cite{luo2020extended} & 72.66\% \\ \hline 
Second-order Hyperbolic CNN \cite{ruthotto2019deep} & 74.3\% \\ \hline 
SOPCNN (RA) \cite{NEURIPS20_FixMatch} &  88.08\%	\\ \hline 
FixMatch  \cite{NEURIPS20_FixMatch} &  89.59\%	\\ \hline 
SESN \cite{sosnovik2019scale} &  91.49\%	\\ \hline 
NSGANetV2 \cite{lu2020nsganetv2} &  92\%	\\ \hline 
FixMatch (RA) \cite{NEURIPS20_FixMatch} &  92.02\%	\\ \hline 
\textbf{GCCN} & \textbf{95.41\%}\\ \hline
\end{tabular}
\end{table}

\paragraph{SVHN}
Table \ref{SVHN} compares the accuracy of GCCN and state-of-the-art methods using the SVHN dataset. GCCN comes slightly after the ReNet+GRU \cite{Moser2020DartsReNet}, FPID \cite{pmlr-v80-hoffman18a} and SE-b \cite{french2017self}. However, GCCN outperformed the accuracy of ReNet+LSTM that processes the image as a sequence of patches. GCCN also outperformed other recent studies, such as DenseNet \cite{huang2017densely} with 94.19, WRN \cite{zagoruyko2016wide} with 94.50\%, E-ABS \cite{ju2020abs} with 89.20\% and  DANN \cite{ganin2016domain} with 91.00\%.

\begin{table}[!ht]
\small
\begin{center}
\caption{Classification accuracy (top 1) results on SVHN data.}\label{SVHN}
\small
\begin{tabular}{|p{6cm}|l|l|}
\hline
Model & Test Accuracy  \\ \hline
E-ABS \cite{ju2020abs}                  & 89.20\% \\ \hline
Asymmetric Tri-Training  \cite{saito2017asymmetric} & 90.83\% \\ \hline
DANN \cite{ganin2016domain}             & 91.00\% \\ \hline
Associative Domain Adaptation \cite{haeusser2017associative} & 91.80\% \\ \hline
SE-a \cite{french2017self}              & 91.92\% \\ \hline
CLS-GAN \cite{qi2020loss}               & 94.02\% \\ \hline  
ReNet+LSTM \cite{Moser2020DartsReNet}   & 94.10\% \\ \hline
DenseNet \cite{huang2017densely}        & 94.19\% \\ \hline 
WRN-OE \cite{hendrycks2019deep}         & 94.19\% \\ \hline
WRN \cite{zagoruyko2016wide}            & 94.50\% \\ \hline
DWT-MEC \cite{roy2019unsupervised}      & 94.62\% \\ \hline 
Farhadi et al., \cite{farhadi2019novel} & 94.62\% \\ \hline
\textbf{GCCN}                           & \textbf{94.65\%} \\ \hline
ReNet+GRU \cite{Moser2020DartsReNet}    & 95.16\% \\ \hline
FPID \cite{pmlr-v80-hoffman18a}         & 95.67\% \\ \hline
\end{tabular}
\end{center}
\end{table}

\paragraph{Vector Augmentation and Normalisation.}
Table \ref{gccn_comp} lists the experiment results of using GCCN vector augmentation and normalisation. We tested GCCN on different image sizes (S), using different CNN networks (Efficient-Net and ResNet) and with various GCCN layers (one, two and three). Each experiment was run under three different settings based on the method of concatenating the global context features with the CNN vector. These were test vector augmentation (Aug), normalisation (norm) and augmentation followed by normalisation (A+N), as explained in the methodology section. Using GCCN with ResNet-50 always produced better accuracy than Efficient-Net. Furthermore, GCCN works better when increasing the image crop size from $32$ to $96$ for the CIFAR-10 and SVHN and $96$ to $224$ for the STL-10. This insight highlights the effectiveness of GCCN in capturing better global attention with higher resolutions of the same images. In most cases, using the normalisation after GCCN augmentation improves the classification accuracy. In multiple experiments, the model overfits if only augmentation or normalisation is utilised. However, the combination method does not show any overfitting.

\begin{table*}[!ht]
\centering
\small
\caption{GCCN vector augmentation and normalisation.}
\label{gccn_comp}
\begin{tabular}{|p{4cm}|l|l|l|l|l|c|}
\hline
Dataset                  & Size                & CNN                     & L & Aug & Norm & Aug+Norm \\ \hline
\multirow{9}{*}{CIFAR10} & \multirow{6}{*}{$32\times 32$} & \multirow{3}{*}{Eff.Net} 
                                                                        & 1 & 86.7\%   & 83.09\%  & 84.62\% \\ \cline{4-7} 
                       &                      &                         & 2 & 86.04\% & 81.51\% & 84.54\% \\ \cline{4-7} 
                       &                      &                         & 3 & 85.87\% & 81.86\% & 83.86\% \\ \cline{3-7} 
                       &                      & \multirow{3}{*}{ResNet} & 1 & 87.97\% & 86.32\% & 88.07\% \\ \cline{4-7} 
                       &                      &                         & 2 & 87.81\% & 84.54\% & \textbf{86.86\%} \\ \cline{4-7} 
                       &                      &                         & 3 & 87.82\% & 84.81\% & 86.56\% \\ \cline{2-7} 
                       & \multirow{3}{*}{$96\times 96$}  & \multirow{3}{*}{ResNet} & 1 & 90.15\% & 93.35\% & \textbf{94.60\%} \\ \cline{4-7} 
                       &                      &                         & 2 & 47\%    & 91.65\% & 93.42\% \\ \cline{4-7} 
                       &                      &                         & 3 & 37\%    & 92.54\% & 94.01\% \\ \hline
\multirow{6}{*}{STL10} & \multirow{3}{*}{$96\times 96$}  & \multirow{3}{*}{ResNet} & 1 & 91.65\% & 91.6 \% & \textbf{92.26\%} \\ \cline{4-7} 
                       &                      &                         & 2 & 90.55\% & 91.44\% & 91.73\% \\ \cline{4-7} 
                       &                      &                         & 3 & 91.09\% & 90.97\% & 91.4 \% \\ \cline{2-7} 
                       & \multirow{3}{*}{$224\times 224$} & \multirow{3}{*}{ResNet} & 1 & 59.54\% & 59.07\% & \textbf{95.41\%} \\ \cline{4-7} 
                       &                      &                         & 2 & 94.45\% & 92.81\% & 94.76\% \\ \cline{4-7} 
                       &                      &                         & 3 & 80.87\% & 69.81\% & 95.24\% \\ \hline
\multirow{6}{*}{SVHN}  & \multirow{3}{*}{$32\times 32$}  & \multirow{3}{*}{ResNet} & 1 & \textbf{94.45\%} & 93.66\% & 94.26\% \\ \cline{4-7} 
                       &                      &                         & 2 & 44.03\% & 93.07\% & 93.63\% \\ \cline{4-7} 
                       &                      &                         & 3 & 34.59\% & 93.18\% & 93.89\% \\ \cline{2-7} 
                       & \multirow{3}{*}{$96\times 96$}  & \multirow{3}{*}{ResNet} & 1 & 94.6\%  & 94.61\% & \textbf{94.65\%} \\ \cline{4-7} 
                       &                      &                         & 2 & 47.59\% & 94.3 \% & 93.89\% \\ \cline{4-7} 
                       &                      &                         & 3 & 94.14\% & 93.91\% & 93.74\% \\ \hline
\end{tabular}
\end{table*}

\subsection{Few-Shot Image Classification}
We followed the state-of-the-art episode composition similar to \cite{snell2017prototypical,vinyals2016matching}. We choose a set of $W$ ways or classes and $K$ support shots per class. Specifically, we tested $1-way$ with $1-shot$ and $5-shots$, and $20-shots$ with $1-shot$ and $5-shots$.
We tested both $Cosine$ and $Euclidean$ distance functions similar to the literature \cite{vinyals2016matching,Sachin2017Optimization}. We utilised a CNN encoder function of four convolutional blocks. Each convolutional block contains $64$ convolution $3 \times 3$ filters. It is also normalised by batch normalisation function \cite{ioffe2015batch} and followed by a Relu non-linearity. Each block also has a $2\times 2$ max-pooling layer. We then add our proposed global context convolutional layer.

\paragraph{Few-Shot Datasets.}
We use three benchmark datasets, including Omniglot MiniImageNet and CUB-200. The training sets are randomly divided into training episodes. Each training episode has a support set and a query set..

\begin{itemize}
    \item MiniImageNet dataset \cite{Sachin2017Optimization} has $60,000$ images of $100$ classes ($600$ images each) from the original ImageNet. Each image in the MiniImageNet is $84\times 84$. To compare with the state-of-the-art, we use the standard split of $64$, $16$ and $20$. It is one of the most difficult datasets for few-shot learning.
    \item CUB-200 dataset \cite{wah2011caltech} has 200 categories of birds and has around 6,000 images for training and 6,000 images for testing.
    \item Omniglot dataset \cite{lake2015human} has $1,623$ handwritten characters from $50$ different alphabets. The dataset is augmented with different rotations, having a total of $6,492$ classes. We used $4,800$ classes of $1200$ characters for training and $1692$ classes for testing, following \cite{vinyals2016matching}. 
\end{itemize}

\paragraph{MiniImageNet.}
Table \ref{tab:mI5w} shows benchmarking results using the MiniImageNet dataset for $5-ways$ with $1-shot$ and $5-shot$. \textit{GCCN} ranks first with $84.8\%$ in the $5-shot$ and second in $1-shot$ with $65.6\%$, after the LaplacianShot, outperforming the state-of-the-art. The LaplacianShot \cite{ziko2020laplacian} has achieved $75.57\%$ and $84.7\%$ in the $1-shot$ and $5-shot$ setups, respectively. Hyperbolic prototypical is a similar work that proposed a new method for the prototypical networks \cite{khrulkov2020hyperbolic}. GCCN outperforms the hyperbolic prototypical with $6\%$ and $8\%$ in $1-shot$ and $5-shot$, respectively. 
Other recent works have also come after \textit{GCCN}, such as Meta-SGD \cite{Li2017MetaSGDLT}, iMAML HF \cite{rajeswaran2019meta}, Meta-Net \cite{munkhdalaiY17Meta} and iMAML GD \cite{rajeswaran2019meta} with $50.5\%$, $49.3\%$, $49.2\%$ and $49\%$, respectively. $GCCN$ outperforms the prototypical and matching networks that achieved $48\%$ and $43.4\%$, respectively. These results show the impact of using the augmented vectors over the conventional CNN vector embeddings. The proposed vector-level augmentation algorithm increased the accuracy of the prototypical networks from $48\%$ to $53\%$, and from $66.2\%$ to $84.8\%$ in the $5-shot$ task. \textit{GCCN} has also outperformed state-of-the-art methods in the $1-shot$ task, such as Meta-learner LSTM, MAML \cite{finn2017model} and GPNet + Polynomial \cite{patacchiola2019deep}.

\begin{table}[!ht]
    \centering
    \small
    \caption{Benchmark of GCCN on MiniImageNet few-shot.}
    \label{tab:mI5w}
    \begin{tabular}{|p{5cm}|l|l|l|}
    \hline
        Model & 1-shot & 5-shot \\ \hline
        PIXELS (Cosine)                                             & 23  \% & 26.6\% \\\hline
        Baseline nearest neighbours (Cosine)                        & 28.8\% & 49.8\% \\\hline
        Matching networks (Cosine) \cite{vinyals2016matching}       & 43.4\% & 51\% \\\hline
        Meta-learner LSTM \cite{Sachin2017Optimization}             & 43.4\% & 60.6\% \\ \hline
        ProtoNet \cite{snell2017prototypical} (Euclid)              & 48\%   & 66.2\% \\ \hline
        MAML \cite{finn2017model}                                   & 48.7\% & 63.1\% \\\hline
        Hyperbolic ProtoNet \cite{khrulkov2020hyperbolic}           & 51.6\% & 66  \% \\\hline
        
        Reptile + Trans \cite{nichol2018first}               & 49.9\% & 65.9\% \\ \hline
        Relation Net \cite{Sung_2018_CVPR}                          & 50.4\% & 65.32\% \\ \hline 
        DN4 \cite{Li_2019_CVPR}                                     & 51.2\% & 71.02\% \\ \hline 
        VAMPIRE \cite{nguyen2020uncertainty}                        & 51.5\% & 64.31\% \\ \hline 
        Hyperbolic ProtoNet (4 Conv) \cite{khrulkov2020hyperbolic}  & 54.43\%& 72.67\% \\ \hline 
        Hyperbolic ProtoNet (ResNet) \cite{khrulkov2020hyperbolic}  & 59.47\%& 76.84\% \\ \hline 
        SImPa (4 Conv)  \cite{nguyen2020pac}                        & 52.1\% & 63.87\% \\ \hline 
        SImPa (ResNet)  \cite{nguyen2020pac}                        &  63.8\%& 78.04\% \\ \hline 
        MAML++ \cite{antoniou2019train}                             & 52.4\% & 68.32\% \\ \hline 
        
        DSN-MR (4 Conv) \cite{simon2020adaptive}                    & 55.88\%& 70.5\% \\ \hline 
        DSN-MR (ResNet) \cite{simon2020adaptive}                    & 64.60\%& 79.51\% \\ \hline 
        EPNet (ResNet) \cite{rodriguez2020embedding}                & 66.50\%& 81.06\% \\ \hline 
        LaplacianShot \cite{ziko2020laplacian}                      & \textbf{75.57\%} & 84.7\% \\ \hline 

        GCCN (ours)                                                 & 65.6\% & \textbf{84.8\%} \\\hline
    \end{tabular}
\end{table}

\paragraph{CUB-200-2011.}
Table \ref{tab:CUB} shows benchmarking results comparing \textit{GCCN} with the state-of-the-art methods using the CUB-200-2011 dataset. GCCN outperformed state-of-the-art methods in both $1-shot$ and $5-shot$. GCCN has $1.6\%$ and $8.56\%$ better accuracy than the hyperbolic prototypical. GCCN also outperforms MAML and MAML++. GCCN has improved the accuracy of the baseline prototypical network. It has $51.31\%$ and $70.77\%$ with around $14\%$ and $10\%$ lower accuracy than GCCN.  

\begin{table}[!ht]
\centering
\small
\caption{5-ways benchmark results on the CUB datasets.}
\label{tab:CUB}
\begin{tabular}{|p{5cm}|l|l|}
\hline
Method                                                     & 1-shot & 5-shot \\ \hline
Matching Nets \cite{vinyals2016matching}                   & 56.53\%         & 63.54\%    \\ \hline
DEML+Matching Nets \cite{vinyals2016matching,zhou2018deep} & 63.47\%         & 64.86\%    \\ \hline
MAML \cite{finn2017model}                                  & 50.45\%         & 59.60\%    \\ \hline
DEML+MAML \cite{zhou2018deep}                              & 64.63\%         & 66.75\%    \\ \hline
Meta-SGD \cite{li2017meta}                                 & 53.34\%         & 67.59\%    \\ \hline
DEML+Meta-SGD \cite{li2017meta,zhou2018deep}               & \textbf{66.95\%}         & 77.11\%    \\ \hline
Hyperbolic ProtoNet \cite{khrulkov2020hyperbolic}          & 64.02\%         & 72.22\%    \\ \hline
ProtoNet \cite{snell2017prototypical}                      & 51.31\%         & 70.77\%    \\ \hline
MACO \cite{hilliard2018few}                                & 60.76\%         & 74.96\%    \\ \hline
RelationNet \cite{Sung_2018_CVPR}                          & 62.45\%         & 76.11\%    \\ \hline
Baseline++ \cite{chen2018a}                                & 60.53\%         & 79.34\%    \\ \hline
GCCN (ours)                                                & 65.62\%         & \textbf{80.74\%}    \\ \hline
\end{tabular}
\end{table}

\paragraph{Omniglot.}
Table \ref{tab:Og520w} shows the benchmarking results comparing \textit{GCCN} with the state-of-the-art and baseline methods using the Omniglot dataset. The results are from the experimental setup of $5-ways$ for both $1-shot$ and $5-shot$ learning. GCCN achieves state-of-the-art results with $99.4\%$ and $99.9\%$ for both $1-shot$ and $5-shot$ tasks. \textit{GCCN} outperformed the original version of the utilised head models (prototypical and matching networks). The prototypical network achieved $98.8\%$ and $99.7\%$, and matching networks have $98.1\%$ and $98.9\%$ in $1-shot$ and $5-shot$ tasks, respectively. Hyperbolic prototypical networks\cite{khrulkov2020hyperbolic} have recently been published to enhance the prototypical networks, with $99\%$ and $99.4\%$. \textit{GCCN} outperformed the hyperbolic version in the $1-shot$ and $5-shot$ learning.
\textit{GCCN} outperforms recent works such VAMPIRE (WACV, 2020) \cite{nguyen2020uncertainty}, APL (ICLR, 2019) \cite{ramalho2019adaptive} and hyperbolic prototypical networks (CVPR, 2020) \cite{khrulkov2020hyperbolic}. \textit{GCCN} using the $Euclidean$ method has the best results over the $Cosine$ and other state-of-the-art networks. \textit{GCCN} proved its superiority in increasing the accuracy of the utilised head models (prototypical and matching networks) using $Euclidean$ or $Cosine$ distance metrics. For example, \textit{GCCN} has increased the prototypical network accuracy from $90\%$ to $99.2$ in the $1-shot$ using $Cosine$.

Table \ref{tab:Og520w} lists benchmarking results of $20-ways$ with both $1-shot$ and $5-shot$ using Omniglot. \textit{GCCN} ranks first in comparison to the state-of-the-art results with $99.1\%$ in the $5-shot$ learning task and second in the $2-shot$ tasks with $96.4\%$, behind the APL \cite{ramalho2019adaptive} that achieved first place with $97.2\%$. The utilised head models (prototypical and matching networks) achieved $95.8\%$ and $98.9\%$, and $62.6\%$ and $74.3\%$, in the $1-shot$ and $5-shots$ tasks, respectively. \textit{GCCN} outperformed both prototypical and matching networks model in the $1-shot$ and $5-shot$ setups. 
Hyperbolic prototypical networks \cite{khrulkov2020hyperbolic} are also outperformed by \textit{GCCN}, achieving $95\%$ and $98.2\%$. \textit{GCCN} also outperforms other state-of-the-art woks, such as VAMPIRE \cite{nguyen2020uncertainty}, adaCNN \cite{munkhdalai2018rapid}, APL \cite{ramalho2019adaptive} and Reptile + Transduction \cite{nichol2018first}. \textit{GCCN} using the $Euclidean$ method has the best results over the $Cosine$ and other state-of-the-art networks. It achieves $21.1\%$ and $12.6\%$ higher scores in the $1-shot$ and $5-shot$ of the $20-ways$, respectively.

\begin{table*}[!ht]
\centering
\small
\caption{Benchmark results on Omniglot on 5-way and 20-way.}
\label{tab:Og520w}
\begin{tabular}{|p{7cm}|p{1cm}|p{1cm}|p{1cm}|p{1cm}|}
\hline
\multirow{2}{*}{Model}                        & \multicolumn{2}{l|}{5-way} & \multicolumn{2}{l|}{20-way} \\ \cline{2-5} 
                                              & 1-shot  & 5-shot           & 1-shot       & 5-shot       \\ \hline
Matching Net \cite{vinyals2016matching} (Cos.)          & 86.4\%          & 90.6\%          & 62.6\%          & 74.3\%        \\ \hline
PrototNet \cite{snell2017prototypical} (Cos.)    & 90\%            & 91\%            & 68.8\%          & 80.1\%        \\ \hline
Reptile + Trans \cite{nichol2018first} & 97.7\%  & 99.5\%           & 89.4\%       & 97\%         \\ \hline
\textbf{APL} \cite{ramalho2019adaptive}                      & 97.9\%          & \textbf{99.9\%} & \textbf{97.2\%} & 97.6\%        \\ \hline
Matching Net \cite{vinyals2016matching} (Euclid.)       & 98.1\%          & 98.9\%          & 92.8\%          & 97.8\%        \\ \hline
Neural statistician \cite{edwards2016towards} & 98.1\%  & 99.5\%           & 93.2\%       & 98.1\%       \\ \hline
VAMPIRE \cite{nguyen2020uncertainty}          & 98.4\%  & 99.6\%           & 93.2\%       & 98.5\%       \\ \hline
adaCNN \cite{munkhdalai2018rapid}             & 98.4\%  & 99.4\%           & 96.1\%       & 98.4\%       \\ \hline
\textbf{MAML} \cite{finn2017model}            & 98.7\%  & \textbf{99.9\%}  & 95.8\%       & 98.9\%       \\ \hline
PrototNet \cite{snell2017prototypical} (Euclid.) & 98.8\%          & 99.7\%          & 95.8\%          & 98.6\%        \\ \hline
Hyperbolic ProtoNet \cite{khrulkov2020hyperbolic}            & 99.0\%          & 99.4\%          & 95.9\%          & 98.2\%        \\ \hline
GCCN (ours) (Cos.)                            & 99.2\%  & 97\%             & 75.3\%       & 86.5\%       \\ \hline
\textbf{GCCN (ours)} (Euclid.)                               & \textbf{99.4\%} & \textbf{99.9\%} & 96.4\%          & \textbf{99.1} \\ \hline
\end{tabular}
\end{table*}

\begin{table}[!ht]
\centering
\small
\caption{Impact of \textit{GCCN} on the prototypical networks.}
\label{PN}
\begin{tabular}{|l|l|l|l|l|}
\hline
distance                   & K-ways & N-shots & Base Model    & Accuracy        \\ \hline
\multirow{8}{*}{Euclidean} & 5      & 1       & \textbf{GCCN} & \textbf{99.4\%} \\ \cline{2-5} 
                           & 5      & 1       & Conv.         & 98.2\%          \\ \cline{2-5} 
                           & 5      & 5       & \textbf{GCCN} & \textbf{99.9\%} \\ \cline{2-5} 
                           & 5      & 5       & Conv.         & 99.4\%          \\ \cline{2-5} 
                           & 20     & 1       & GCCN          & \textbf{96.4\%} \\ \cline{2-5} 
                           & 20     & 1       & Conv.         & 95.8\%          \\ \cline{2-5} 
                           & 20     & 5       & \textbf{GCCN} & \textbf{99.1\%} \\ \cline{2-5} 
                           & 20     & 5       & Conv.         & 98.6\%          \\ \hline
\multirow{8}{*}{Cosine}    & 5      & 1       & \textbf{GCCN} & \textbf{99.2\%} \\ \cline{2-5} 
                           & 5      & 1       & Conv.         & 90\%            \\ \cline{2-5} 
                           & 5      & 5       & \textbf{GCCN} & \textbf{97\%}   \\ \cline{2-5} 
                           & 5      & 5       & Conv.         & 91\%            \\ \cline{2-5} 
                           & 20     & 1       & \textbf{GCCN} & \textbf{75.3\%} \\ \cline{2-5} 
                           & 20     & 1       & Conv.         & 68.8\%          \\ \cline{2-5} 
                           & 20     & 5       & \textbf{GCCN} & \textbf{86.5\%} \\ \cline{2-5} 
                           & 20     & 5       & Conv.         & 80.1\%          \\ \hline
\end{tabular}
\end{table}

\begin{table}[!ht]
\centering
\small
\caption{Impact of \textit{GCCN} on the matching networks.}
\label{MN}
\begin{tabular}{|l|l|l|l|l|}
\hline
distance                   & K-ways & N-shots & Base Model    & Accuracy        \\ \hline
\multirow{8}{*}{Euclidean} & 5      & 1       & \textbf{GCCN} & \textbf{98.8\%} \\ \cline{2-5} 
                           & 5      & 1       & Conv.         & 98.1\%          \\ \cline{2-5} 
                           & 5      & 5       & \textbf{GCCN} & \textbf{99.9\%} \\ \cline{2-5} 
                           & 5      & 5       & Conv.         & 98.9\%          \\ \cline{2-5} 
                           & 20     & 1       & GCCN          & 91.7\%          \\ \cline{2-5} 
                           & 20     & 1       & Conv.         & 92.8\%          \\ \cline{2-5} 
                           & 20     & 5       & \textbf{GCCN} & \textbf{98.7\%} \\ \cline{2-5} 
                           & 20     & 5       & Conv.         & 97.8\%          \\ \hline
\multirow{8}{*}{Cosine}    & 5      & 1       & \textbf{GCCN} & \textbf{89.0\%} \\ \cline{2-5} 
                           & 5      & 1       & Conv.         & 86.4\%          \\ \cline{2-5} 
                           & 5      & 5       & \textbf{GCCN} & \textbf{95.8\%} \\ \cline{2-5} 
                           & 5      & 5       & Conv.         & 90.6\%          \\ \cline{2-5} 
                           & 20     & 1       & \textbf{GCCN} & \textbf{71.4\%} \\ \cline{2-5} 
                           & 20     & 1       & Conv.         & 62.6\%          \\ \cline{2-5} 
                           & 20     & 5       & \textbf{GCCN} & \textbf{77.7\%} \\ \cline{2-5} 
                           & 20     & 5       & Conv.         & 74.3\%          \\ \hline
\end{tabular}
\end{table}

\paragraph{GCCN Impact on Prototypical and Matching Networks.}
In this section, we discuss the impact of the proposed \textit{GCCN} used as a base model within the prototypical and matching networks. Tables \ref{PN} and \ref{MN} list the experiment results using both the $Euclidean$ and $Cosine$ distance measures with the prototypical and matching networks, respectively. 

Prototypical networks using the proposed vector augmentation algorithm (\textit{GCCN}) have better accuracy than their original implementation using the classical convolutional networks. The vector augmentation method has successfully enriched the CNN feature embeddings. The \textit{GCCN} performance is better in both $Euclidean$ distance and $Cosine$ similarity measures. The performance of prototypical networks using \textit{GCCN} has outperformed the original prototypical networks in all tasks. For example, \textit{GCCN} using the $Cosine$ method has significantly improved the accuracy from $90\%$, $91\%$, $68.8\%$ and $80.1\%$ to $99.2\%$, $97\%$, $75.3\%$ and $86.5\%$ in the tasks of $5-ways-1-shot$, $5-ways-5-shots$, $20-ways-1-shot$ and $20-ways-5-shots$, respectively. The performance has also been improved using the $Euclidean$ distance in all tasks.

We also implemented the matching networks using the proposed \textit{GCCN} as a base model instead of CNN. Matching networks with \textit{GCCN} outperform the original version of matching with the CNN base model. The matching networks ($Cosine$) performance is significantly improved using the proposed vector embedding augmentation utilising global context structural information. Table \ref{MN} shows that the accuracy of matching networks increased from $90.6\%$ to $95.8\%$ for the $5-ways-5-shots$ task and $62.6\%$ to $71.4\%$ for the $20-ways-1-shots$ task. On the other hand, the matching networks ($Euclidean$) performance also improved. Matching networks with \textit{GCCN} outperformed the original version in most cases. For example, the accuracy increased from $98.1\%$, $98.9\%$ and $97.8\%$ to $98.8\%$, $99.9\%$ and $98.7\%$ in the tasks of $5-ways-1-shots$, $5-ways-5-shots$, and $20-ways-5-shots$, respectively. 

\section{Conclusion}
We have introduced \textit{GCCN}, a novel embedding vector augmentation and normalisation method. The proposed \textit{GCCN} tends to overcome the limitation of the traditional CNNs of ignoring important structural information relying on local receptive fields. We offer to extract useful global context information to augment the CNN features. This augmentation has proved to be a simple yet effective approach. In this paper, we have experimented with this methodology on both image classification and few-shot learning datasets. We have also introduced an in-depth performance evaluation using the proposed vector embedding method under the state-of-the-art few-shot methods.

\section*{Acknowledgments}
Ali Hamdi is supported by RMIT Research Stipend Scholarship. 
This research is partially supported by Australian Research Council (ARC) Discovery Project \textit{DP190101485}.
\bibliographystyle{elsarticle-num}
\bibliography{bib}

\begin{thebibliography}{10}
\expandafter\ifx\csname url\endcsname\relax
  \def\url#1{\texttt{#1}}\fi
\expandafter\ifx\csname urlprefix\endcsname\relax\def\urlprefix{URL }\fi
\expandafter\ifx\csname href\endcsname\relax
  \def\href#1#2{#2} \def\path#1{#1}\fi

\bibitem{luo2016understanding}
W.~Luo, Y.~Li, R.~Urtasun, R.~Zemel, Understanding the effective receptive
  field in deep convolutional neural networks, in: Proceedings of the 30th
  International Conference on Neural Information Processing Systems, 2016, pp.
  4905--4913.

\bibitem{Liu_2021_CVPR}
J.~Liu, C.~Li, F.~Liang, C.~Lin, M.~Sun, J.~Yan, W.~Ouyang, D.~Xu, Inception
  convolution with efficient dilation search, in: Proceedings of the IEEE/CVF
  Conference on Computer Vision and Pattern Recognition (CVPR), 2021, pp.
  11486--11495.

\bibitem{szegedy2015going}
C.~Szegedy, W.~Liu, Y.~Jia, P.~Sermanet, S.~Reed, D.~Anguelov, D.~Erhan,
  V.~Vanhoucke, A.~Rabinovich, Going deeper with convolutions, in: Proceedings
  of the IEEE conference on computer vision and pattern recognition, 2015, pp.
  1--9.

\bibitem{he2016deep}
K.~He, X.~Zhang, S.~Ren, J.~Sun, Deep residual learning for image recognition,
  in: Proceedings of the IEEE conference on computer vision and pattern
  recognition, 2016, pp. 770--778.

\bibitem{huang2017densely}
G.~Huang, Z.~Liu, L.~Van Der~Maaten, K.~Q. Weinberger, Densely connected
  convolutional networks, in: Proceedings of the IEEE conference on computer
  vision and pattern recognition, 2017, pp. 4700--4708.

\bibitem{tan2019efficientnet}
M.~Tan, Q.~Le, Efficientnet: Rethinking model scaling for convolutional neural
  networks, in: International Conference on Machine Learning, PMLR, 2019, pp.
  6105--6114.

\bibitem{masi2016we}
I.~Masi, A.~T. Tran, T.~Hassner, J.~T. Leksut, G.~Medioni, Do we really need to
  collect millions of faces for effective face recognition?, in: European
  Conference on Computer Vision, Springer, 2016, pp. 579--596.

\bibitem{zhou2018graph}
J.~Zhou, G.~Cui, Z.~Zhang, C.~Yang, Z.~Liu, M.~Sun, Graph neural networks: A
  review of methods and applications, arXiv preprint arXiv:1812.08434 (2018).

\bibitem{gao2019graph}
J.~Gao, T.~Zhang, C.~Xu, Graph convolutional tracking, in: Proceedings of the
  IEEE Conference on Computer Vision and Pattern Recognition, 2019, pp.
  4649--4659.

\bibitem{hamdi2020flexgrid2vec}
A.~Hamdi, D.~Y. Kim, F.~Salim, flexgrid2vec: Learning efficient visual
  representations vectors, arXiv e-prints (2020) arXiv--2007.

\bibitem{vinyals2016matching}
O.~Vinyals, C.~Blundell, T.~Lillicrap, D.~Wierstra, et~al., Matching networks
  for one shot learning, in: Advances in neural information processing systems,
  2016, pp. 3630--3638.

\bibitem{snell2017prototypical}
J.~Snell, K.~Swersky, R.~Zemel, Prototypical networks for few-shot learning,
  in: Advances in neural information processing systems, 2017, pp. 4077--4087.

\bibitem{sung2018learning}
F.~Sung, Y.~Yang, L.~Zhang, T.~Xiang, P.~H. Torr, T.~M. Hospedales, Learning to
  compare: Relation network for few-shot learning, in: Proceedings of the IEEE
  Conference on Computer Vision and Pattern Recognition, 2018, pp. 1199--1208.

\bibitem{ioffe2015batch}
S.~Ioffe, C.~Szegedy, Batch normalization: Accelerating deep network training
  by reducing internal covariate shift, arXiv preprint arXiv:1502.03167 (2015).

\bibitem{ba2016layer}
J.~L. Ba, J.~R. Kiros, G.~E. Hinton, Layer normalization, arXiv preprint
  arXiv:1607.06450 (2016).

\bibitem{antoniou2017data}
A.~Antoniou, A.~Storkey, H.~Edwards, Data augmentation generative adversarial
  networks, arXiv preprint arXiv:1711.04340 (2017).

\bibitem{mensink2013distance}
T.~Mensink, J.~Verbeek, F.~Perronnin, G.~Csurka, Distance-based image
  classification: Generalizing to new classes at near-zero cost, IEEE
  transactions on pattern analysis and machine intelligence 35~(11) (2013)
  2624--2637.

\bibitem{rippel2015metric}
O.~Rippel, M.~Paluri, P.~Dollar, L.~Bourdev, Metric learning with adaptive
  density discrimination, arXiv preprint arXiv:1511.05939 (2015).

\bibitem{nguyen2020uncertainty}
C.~Nguyen, T.-T. Do, G.~Carneiro, Uncertainty in model-agnostic meta-learning
  using variational inference, in: The IEEE Winter Conference on Applications
  of Computer Vision, 2020, pp. 3090--3100.

\bibitem{ramalho2019adaptive}
T.~Ramalho, M.~Garnelo, Adaptive posterior learning: few-shot learning with a
  surprise-based memory module, arXiv preprint arXiv:1902.02527 (2019).

\bibitem{nguyen2020pac}
C.~Nguyen, T.-T. Do, G.~Carneiro, Pac-bayesian meta-learning with implicit
  prior, IEEE TRANSACTION ON PATTERN ANALYSIS AND MACHINE INTELLIGENCE (2020).

\bibitem{ziko2020laplacian}
I.~Ziko, J.~Dolz, E.~Granger, I.~B. Ayed, Laplacian regularized few-shot
  learning, in: International Conference on Machine Learning, PMLR, 2020, pp.
  11660--11670.

\bibitem{khrulkov2020hyperbolic}
V.~Khrulkov, L.~Mirvakhabova, E.~Ustinova, I.~Oseledets, V.~Lempitsky,
  Hyperbolic image embeddings, in: Proceedings of the IEEE/CVF Conference on
  Computer Vision and Pattern Recognition, 2020, pp. 6418--6428.

\bibitem{huang2016deep}
G.~Huang, Y.~Sun, Z.~Liu, D.~Sedra, K.~Q. Weinberger, Deep networks with
  stochastic depth, in: European conference on computer vision, Springer, 2016,
  pp. 646--661.

\bibitem{he2016identity}
K.~He, X.~Zhang, S.~Ren, J.~Sun, Identity mappings in deep residual networks,
  in: European conference on computer vision, Springer, 2016, pp. 630--645.

\bibitem{chen2019closer}
W.-Y. Chen, Y.-C. Liu, Z.~Kira, Y.-C.~F. Wang, J.-B. Huang, A closer look at
  few-shot classification, arXiv preprint arXiv:1904.04232 (2019).

\bibitem{finn2017model}
C.~Finn, P.~Abbeel, S.~Levine, Model-agnostic meta-learning for fast adaptation
  of deep networks, arXiv preprint arXiv:1703.03400 (2017).

\bibitem{wang2018low}
Y.-X. Wang, R.~Girshick, M.~Hebert, B.~Hariharan, Low-shot learning from
  imaginary data, in: Proceedings of the IEEE conference on computer vision and
  pattern recognition, 2018, pp. 7278--7286.

\bibitem{Sachin2017Optimization}
S.~Ravi, H.~Larochelle,
  \href{https://openreview.net/forum?id=rJY0-Kcll}{Optimization as a model for
  few-shot learning}, in: 5th International Conference on Learning
  Representations, {ICLR} 2017, Toulon, France, April 24-26, 2017, Conference
  Track Proceedings, OpenReview.net, 2017.
\newline\urlprefix\url{https://openreview.net/forum?id=rJY0-Kcll}

\bibitem{antoniou2019how}
A.~Antoniou, H.~Edwards, A.~J. Storkey,
  \href{https://openreview.net/forum?id=HJGven05Y7}{How to train your {MAML}},
  in: 7th International Conference on Learning Representations, {ICLR} 2019,
  New Orleans, LA, USA, May 6-9, 2019, OpenReview.net, 2019.
\newline\urlprefix\url{https://openreview.net/forum?id=HJGven05Y7}

\bibitem{krizhevsky2009learning}
A.~Krizhevsky, G.~Hinton, et~al., Learning multiple layers of features from
  tiny images (2009).

\bibitem{coates2011analysis}
A.~Coates, A.~Ng, H.~Lee, An analysis of single-layer networks in unsupervised
  feature learning, in: Proceedings of the fourteenth international conference
  on artificial intelligence and statistics, 2011, pp. 215--223.

\bibitem{hjelm2018learning}
R.~D. Hjelm, A.~Fedorov, S.~Lavoie-Marchildon, K.~Grewal, P.~Bachman,
  A.~Trischler, Y.~Bengio, Learning deep representations by mutual information
  estimation and maximization, arXiv preprint arXiv:1808.06670 (2018).

\bibitem{NIPS2019_8577}
E.~Dupont, A.~Doucet, Y.~W. Teh, Augmented neural odes, in: H.~Wallach,
  H.~Larochelle, A.~Beygelzimer, F.~d\textquotesingle Alch\'{e}-Buc, E.~Fox,
  R.~Garnett (Eds.), Advances in Neural Information Processing Systems 32,
  Curran Associates, Inc., 2019, pp. 3140--3150.

\bibitem{qi2020loss}
G.-J. Qi, Loss-sensitive generative adversarial networks on lipschitz
  densities, International Journal of Computer Vision 128~(5) (2020)
  1118--1140.

\bibitem{misra2020mish}
D.~Misra, Mish: A self regularized non-monotonic neural activation function,
  in: British Machine Vision Conference (BMVC), 2020.

\bibitem{radford2015unsupervised}
A.~Radford, L.~Metz, S.~Chintala, Unsupervised representation learning with
  deep convolutional generative adversarial networks, arXiv preprint
  arXiv:1511.06434 (2015).

\bibitem{gonzalez2020improved}
S.~Gonzalez, R.~Miikkulainen, Improved training speed, accuracy, and data
  utilization through loss function optimization, in: 2020 IEEE Congress on
  Evolutionary Computation (CEC), IEEE, 2020, pp. 1--8.

\bibitem{oyallon2017scaling}
E.~Oyallon, E.~Belilovsky, S.~Zagoruyko, Scaling the scattering transform: Deep
  hybrid networks, in: Proceedings of the IEEE international conference on
  computer vision, 2017, pp. 5618--5627.

\bibitem{sabour2017dynamic}
S.~Sabour, N.~Frosst, G.~E. Hinton, Dynamic routing between capsules, in:
  Advances in neural information processing systems, 2017, pp. 3856--3866.

\bibitem{hendrycks2016baseline}
D.~Hendrycks, K.~Gimpel, A baseline for detecting misclassified and
  out-of-distribution examples in neural networks, in: International Conference
  on Learning Representations (ICLR), 2017.

\bibitem{sato2015apac}
I.~Sato, H.~Nishimura, K.~Yokoi, Apac: Augmented pattern classification with
  neural networks, arXiv preprint arXiv:1505.03229 (2015).

\bibitem{liao2016importance}
Z.~Liao, G.~Carneiro, On the importance of normalisation layers in deep
  learning with piecewise linear activation units, in: 2016 IEEE Winter
  Conference on Applications of Computer Vision (WACV), IEEE, 2016, pp. 1--8.

\bibitem{lee2015deeply}
C.-Y. Lee, S.~Xie, P.~Gallagher, Z.~Zhang, Z.~Tu, Deeply-supervised nets, in:
  Artificial intelligence and statistics, PMLR, 2015, pp. 562--570.

\bibitem{courbariaux2015binaryconnect}
M.~Courbariaux, Y.~Bengio, J.-P. David, Binaryconnect: Training deep neural
  networks with binary weights during propagations, in: NIPS, 2015.

\bibitem{NEURIPS2019_MixMatch}
D.~Berthelot, N.~Carlini, I.~Goodfellow, N.~Papernot, A.~Oliver, C.~A. Raffel,
  Mixmatch: A holistic approach to semi-supervised learning, in: Advances in
  Neural Information Processing Systems, Vol.~32, Curran Associates, Inc.,
  2019.

\bibitem{huang2020dianet}
Z.~Huang, S.~Liang, M.~Liang, H.~Yang, Dianet: Dense-and-implicit attention
  network, in: Proceedings of the AAAI Conference on Artificial Intelligence,
  Vol.~34, 2020, pp. 4206--4214.

\bibitem{clevert2015fast}
D.-A. Clevert, T.~Unterthiner, S.~Hochreiter, Fast and accurate deep network
  learning by exponential linear units (elus), arXiv preprint arXiv:1511.07289
  (2015).

\bibitem{real2017large}
E.~Real, S.~Moore, A.~Selle, S.~Saxena, Y.~L. Suematsu, J.~Tan, Q.~V. Le,
  A.~Kurakin, Large-scale evolution of image classifiers, in: International
  Conference on Machine Learning, PMLR, 2017, pp. 2902--2911.

\bibitem{luo2020extended}
C.~Luo, J.~Zhan, L.~Wang, W.~Gao, Extended batch normalization, arXiv preprint
  arXiv:2003.05569 (2020).

\bibitem{ruthotto2019deep}
L.~Ruthotto, E.~Haber, Deep neural networks motivated by partial differential
  equations, Journal of Mathematical Imaging and Vision (2019) 1--13.

\bibitem{NEURIPS20_FixMatch}
K.~Sohn, D.~Berthelot, N.~Carlini, Z.~Zhang, H.~Zhang, C.~A. Raffel, E.~D.
  Cubuk, A.~Kurakin, C.-L. Li, Fixmatch: Simplifying semi-supervised learning
  with consistency and confidence, in: H.~Larochelle, M.~Ranzato, R.~Hadsell,
  M.~F. Balcan, H.~Lin (Eds.), Advances in Neural Information Processing
  Systems, Vol.~33, Curran Associates, Inc., 2020, pp. 596--608.

\bibitem{sosnovik2019scale}
I.~Sosnovik, M.~Szmaja, A.~Smeulders, Scale-equivariant steerable networks, in:
  International Conference on Learning Representations, 2019.

\bibitem{lu2020nsganetv2}
Z.~Lu, K.~Deb, E.~Goodman, W.~Banzhaf, V.~N. Boddeti, Nsganetv2: Evolutionary
  multi-objective surrogate-assisted neural architecture search, in: European
  Conference on Computer Vision, Springer, 2020, pp. 35--51.

\bibitem{Moser2020DartsReNet}
B.~B. Moser, F.~Raue, J.~Hees, A.~Dengel, Dartsrenet: Exploring new rnn cells
  in renet architectures, in: I.~Farka{\v{s}}, P.~Masulli, S.~Wermter (Eds.),
  Artificial Neural Networks and Machine Learning -- ICANN 2020, Springer
  International Publishing, Cham, 2020, pp. 850--861.

\bibitem{pmlr-v80-hoffman18a}
J.~Hoffman, E.~Tzeng, T.~Park, J.-Y. Zhu, P.~Isola, K.~Saenko, A.~Efros,
  T.~Darrell, Cycada: Cycle-consistent adversarial domain adaptation, in:
  International conference on machine learning, PMLR, 2018, pp. 1989--1998.

\bibitem{french2017self}
G.~French, M.~Mackiewicz, M.~Fisher, Self-ensembling for visual domain
  adaptation, arXiv preprint arXiv:1706.05208 (2017).

\bibitem{zagoruyko2016wide}
S.~Zagoruyko, N.~Komodakis, Wide residual networks, arXiv preprint
  arXiv:1605.07146 (2016).

\bibitem{ju2020abs}
A.~Ju, D.~Wagner, E-abs: Extending the analysis-by-synthesis robust
  classification model to more complex image domains, in: Proceedings of the
  13th ACM Workshop on Artificial Intelligence and Security, 2020, pp. 25--36.

\bibitem{ganin2016domain}
Y.~Ganin, E.~Ustinova, H.~Ajakan, P.~Germain, H.~Larochelle, F.~Laviolette,
  M.~Marchand, V.~Lempitsky, Domain-adversarial training of neural networks,
  The Journal of Machine Learning Research 17~(1) (2016) 2096--2030.

\bibitem{saito2017asymmetric}
K.~Saito, Y.~Ushiku, T.~Harada, Asymmetric tri-training for unsupervised domain
  adaptation, arXiv preprint arXiv:1702.08400 (2017).

\bibitem{haeusser2017associative}
P.~Haeusser, T.~Frerix, A.~Mordvintsev, D.~Cremers, Associative domain
  adaptation, in: Proceedings of the IEEE International Conference on Computer
  Vision, 2017, pp. 2765--2773.

\bibitem{hendrycks2019deep}
D.~Hendrycks, M.~Mazeika, T.~Dietterich, Deep anomaly detection with outlier
  exposure, in: International Conference on Learning Representations, 2019.

\bibitem{roy2019unsupervised}
S.~Roy, A.~Siarohin, E.~Sangineto, S.~R. Bulo, N.~Sebe, E.~Ricci, Unsupervised
  domain adaptation using feature-whitening and consensus loss, in: Proceedings
  of the IEEE Conference on Computer Vision and Pattern Recognition, 2019, pp.
  9471--9480.

\bibitem{farhadi2019novel}
M.~Farhadi, M.~Ghasemi, Y.~Yang, A novel design of adaptive and hierarchical
  convolutional neural networks using partial reconfiguration on fpga, in: 2019
  IEEE High Performance Extreme Computing Conference (HPEC), IEEE, 2019, pp.
  1--7.

\bibitem{wah2011caltech}
C.~Wah, S.~Branson, P.~Welinder, P.~Perona, S.~Belongie, The caltech-ucsd
  birds-200-2011 dataset (2011).

\bibitem{lake2015human}
B.~M. Lake, R.~Salakhutdinov, J.~B. Tenenbaum, Human-level concept learning
  through probabilistic program induction, Science 350~(6266) (2015)
  1332--1338.

\bibitem{Li2017MetaSGDLT}
Z.~Li, F.~Zhou, F.~Chen, H.~Li, Meta-sgd: Learning to learn quickly for few
  shot learning, ArXiv abs/1707.09835 (2017).

\bibitem{rajeswaran2019meta}
A.~Rajeswaran, C.~Finn, S.~M. Kakade, S.~Levine, Meta-learning with implicit
  gradients, in: Advances in Neural Information Processing Systems, 2019, pp.
  113--124.

\bibitem{munkhdalaiY17Meta}
T.~Munkhdalai, H.~Yu, Meta networks, in: International Conference on Machine
  Learning (ICML), Vol. abs/1703.00837, 2017.

\bibitem{patacchiola2019deep}
M.~Patacchiola, J.~Turner, E.~J. Crowley, M.~O'Boyle, A.~Storkey, Deep kernel
  transfer in gaussian processes for few-shot learning, arXiv preprint
  arXiv:1910.05199 (2019).

\bibitem{nichol2018first}
A.~Nichol, J.~Achiam, J.~Schulman, On first-order meta-learning algorithms,
  arXiv preprint arXiv:1803.02999 (2018).

\bibitem{Sung_2018_CVPR}
F.~Sung, Y.~Yang, L.~Zhang, T.~Xiang, P.~H. Torr, T.~M. Hospedales, Learning to
  compare: Relation network for few-shot learning, in: Proceedings of the IEEE
  Conference on Computer Vision and Pattern Recognition (CVPR), 2018.

\bibitem{Li_2019_CVPR}
W.~Li, L.~Wang, J.~Xu, J.~Huo, Y.~Gao, J.~Luo, Revisiting local descriptor
  based image-to-class measure for few-shot learning, in: Proceedings of the
  IEEE/CVF Conference on Computer Vision and Pattern Recognition (CVPR), 2019.

\bibitem{antoniou2019train}
A.~Antoniou, H.~Edwards, A.~Storkey, How to train your maml, in: International
  Conference on Learning Representations, 2019.

\bibitem{simon2020adaptive}
C.~Simon, P.~Koniusz, R.~Nock, M.~Harandi, Adaptive subspaces for few-shot
  learning, in: Proceedings of the IEEE/CVF Conference on Computer Vision and
  Pattern Recognition, 2020, pp. 4136--4145.

\bibitem{rodriguez2020embedding}
P.~Rodr{\'\i}guez, I.~Laradji, A.~Drouin, A.~Lacoste, Embedding propagation:
  Smoother manifold for few-shot classification, in: European Conference on
  Computer Vision, Springer, 2020, pp. 121--138.

\bibitem{zhou2018deep}
F.~Zhou, B.~Wu, Z.~Li, Deep meta-learning: Learning to learn in the concept
  space, arXiv preprint arXiv:1802.03596 (2018).

\bibitem{li2017meta}
Z.~Li, F.~Zhou, F.~Chen, H.~Li, Meta-sgd: Learning to learn quickly for
  few-shot learning, arXiv preprint arXiv:1707.09835 (2017).

\bibitem{hilliard2018few}
N.~Hilliard, L.~Phillips, S.~Howland, A.~Yankov, C.~D. Corley, N.~O. Hodas,
  Few-shot learning with metric-agnostic conditional embeddings, arXiv preprint
  arXiv:1802.04376 (2018).

\bibitem{chen2018a}
W.-Y. Chen, Y.-C. Liu, Z.~Kira, Y.-C.~F. Wang, J.-B. Huang,
  \href{https://openreview.net/forum?id=HkxLXnAcFQ}{A closer look at few-shot
  classification}, in: International Conference on Learning Representations,
  2019.
\newline\urlprefix\url{https://openreview.net/forum?id=HkxLXnAcFQ}

\bibitem{munkhdalai2018rapid}
T.~Munkhdalai, X.~Yuan, S.~Mehri, A.~Trischler, Rapid adaptation with
  conditionally shifted neurons, in: International Conference on Machine
  Learning, PMLR, 2018, pp. 3664--3673.

\bibitem{edwards2016towards}
H.~Edwards, A.~Storkey, Towards a neural statistician, in: 5th International
  Conference on Learning Representations, {ICLR} 2017, Toulon, France, April
  24-26, 2017, Conference Track Proceedings, OpenReview.net, 2017.

\end{thebibliography}

\end{document}